\documentclass[11pt]{article}

\usepackage{acl}
\usepackage{times}
\usepackage{latexsym}
\usepackage[T1]{fontenc}
\usepackage[utf8]{inputenc}
\usepackage{microtype}
\usepackage{inconsolata}
\usepackage{graphicx}
\usepackage{booktabs}
\usepackage{multirow}
\usepackage{tikz}
\usetikzlibrary{shapes,shapes.geometric,arrows,arrows.meta,positioning,calc,fit}

\title{Reinforced Agent: Inference-Time Feedback for Tool-Calling Agents}

\author{
Anh Ta \quad Junjie Zhu \quad Shahin Shayandeh \\
Apple \\
\texttt{\{atta, jason.zhu, shn\}@apple.com}
}

\date{}

\begin{document}

\maketitle

\begin{abstract}
Tool-calling agents are evaluated on tool selection, parameter accuracy, and scope recognition, yet LLM trajectory assessments remain inherently \textit{post-hoc}. Disconnected from the active execution loop, such assessments identify errors that are usually addressed through prompt-tuning or retraining, and fundamentally cannot course-correct the agent in real time. To close this gap, we move evaluation into the execution loop at \textit{inference time}: a specialized reviewer agent evaluates provisional tool calls \textit{prior to} execution, shifting the paradigm from post-hoc recovery to proactive evaluation and error mitigation.

In practice, this architecture establishes a clear separation of concerns between the primary execution agent and a secondary review agent. As with any multi-agent system, the reviewer can introduce new errors while correcting others, yet no prior work to our knowledge has systematically measured this tradeoff. To quantify this tradeoff, we introduce \textit{Helpfulness-Harmfulness metrics}: helpfulness measures the percentage of base agent errors that feedback corrects; harmfulness measures the percentage of correct responses that feedback degrades. These metrics directly inform reviewer design by revealing whether a given model or prompt provides net positive value.

We evaluate our approach on BFCL (single-turn) and $\tau^2$-Bench (multi-turn stateful scenarios), achieving +5.5\% on irrelevance detection and +7.1\% on multi-turn tasks. Our metrics reveal that reviewer model choice is critical: the reasoning model o3-mini achieves a 3:1 benefit-to-risk ratio versus 2.1:1 for GPT-4o. Automated prompt optimization via GEPA provides an additional +1.5--2.8\%. Together, these results demonstrate a core advantage of separating execution and review: the reviewer can be systematically improved through model selection and prompt optimization, without retraining the base agent.

\end{abstract}

% Introduction
\section{Introduction}

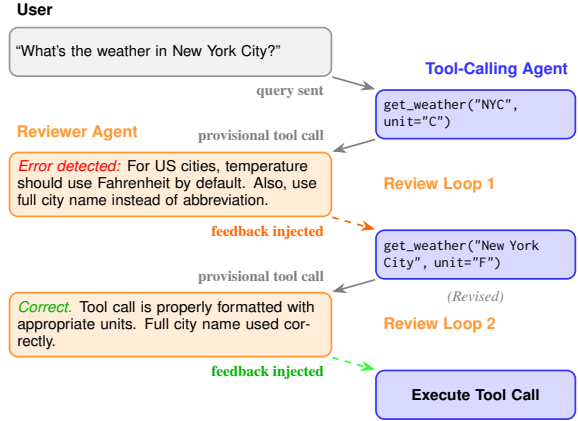
\begin{figure}[t]
    \centering
    \resizebox{\columnwidth}{!}{
        \begin{tikzpicture}[
    node distance=1.2cm and 2cm,
    block/.style={rectangle, rounded corners, draw=black, thick, text width=2.75cm, minimum height=0.8cm, align=left, font=\sffamily\scriptsize},
    user/.style={block, fill=gray!10, draw=gray!80},
    agent/.style={block, fill=blue!15, draw=blue!80},
    feedback/.style={block, fill=orange!15, draw=orange!80},
    execute/.style={block, fill=blue!15, draw=blue!80, text width=2.75cm, minimum height=0.6cm, align=center, font=\sffamily\scriptsize\bfseries},
    arrow/.style={-Stealth, thick, gray},
    label/.style={font=\sffamily\scriptsize\bfseries, text width=1.5cm, align=right},
    entitylabel/.style={font=\sffamily\fontsize{7.5}{9}\selectfont\bfseries},
    % Fixed vertical gap between arrow start and end points
    vgap/.style={yshift=-1.2cm},
    hgap/.style={xshift=2cm}
]

% Define fixed gaps for consistent arrow lengths
\def\arrowgap{0.2cm}      % Vertical gap between bottom of one box and top of next
\def\hcolumngap{0.7cm}    % Horizontal gap between right edge of left boxes and left edge of right boxes
\def\leftboxwidth{5cm} % Width of left column boxes (User, Feedback) - 2.75cm * 1.5
\def\rightboxwidth{3cm} % Width of right column boxes (Tool-Calling Agent, Execute) - 2.75cm * 1.2

% User Query (left side, anchor position)
\node (query) [user, text width=\leftboxwidth, xshift=-3cm] {
    % \textbf{User Query:}
    ``What's the weather in New York City?''
};
% Single label above query, left-aligned
\node[entitylabel, text=black, above=0.05cm of query.north west, anchor=south west] {User};

% Agent 1: Position with fixed gap below query
\node (agent1) [agent, text width=\rightboxwidth, anchor=north west] at ($(query.south east) + (\hcolumngap, -\arrowgap)$) {
    \texttt{get\_weather("NYC", unit="C")}
};
% Single label above agent1, right-aligned
\node[entitylabel, text=blue!80, above=0.05cm of agent1.north east, anchor=south east] {Tool-Calling Agent};

% Feedback 1: Position with fixed gap below agent1
\node (feedback1) [feedback, text width=\leftboxwidth, anchor=north east] at ($(agent1.south west) + (-\hcolumngap, -\arrowgap)$) {
    \textit{\textcolor{red}{Error detected:}} For US cities, temperature should use Fahrenheit by default. Also, use full city name instead of abbreviation.
};
% Single label above feedback1, left-aligned
\node[entitylabel, text=orange!80, above=0.05cm of feedback1.north west, anchor=south west] {Reviewer Agent};

% Loop 1 Label
\node (loop1label) [entitylabel, text=orange!80, anchor=west] at (feedback1.east -| agent1.west) {Review Loop 1};

% Agent 2: Position with fixed gap below feedback1
\node (agent2) [agent, text width=\rightboxwidth, anchor=north west] at ($(feedback1.south east) + (\hcolumngap, -\arrowgap)$) {
    \texttt{get\_weather("New York City", unit="F")}
};
\node[font=\scriptsize\itshape, text=gray, below=0.05cm of agent2] {(Revised)};

% Feedback 2: Position with fixed gap below agent2
\node (feedback2) [feedback, text width=\leftboxwidth, anchor=north east] at ($(agent2.south west) + (-\hcolumngap, -\arrowgap)$) {
    \textit{\textcolor{green!70!black}{Correct.}} Tool call is properly formatted with appropriate units. Full city name used correctly.
};
%\node[label, left=0.2cm of feedback2, text=orange!80] {Feedback\\Agent\\(o3-mini)};

% Loop 2 Label
\node (loop2label) [entitylabel, text=orange!80, anchor=west] at (feedback2.east -| agent2.west) {Review Loop 2};

% Execute: Position with fixed gap below feedback2
\node (execute) [agent, text width=\rightboxwidth, align=center, anchor=north west] at ($(feedback2.south east) + (\hcolumngap, -\arrowgap)$) {
    \textbf{Execute Tool Call}
};

% Arrows (all with same length due to fixed gaps)
% Arrow 1: query to agent1
\draw[arrow] (query.south east) -- (agent1.north west);
\node[font=\scriptsize\bfseries, text=gray, anchor=north east] at (query.south east) {query sent};

% Arrow 2: agent1 to feedback1
\draw[arrow] (agent1.south west) -- (feedback1.north east);
\node[font=\scriptsize\bfseries, text=gray, anchor=south east] at (feedback1.north east) {provisional tool call};

% Arrow 3: feedback1 to agent2 (feedback)
\draw[arrow, dashed, orange!80!red] (feedback1.south east) -- (agent2.north west);
\node[font=\scriptsize\bfseries, text=orange!80!red, anchor=north east] at (feedback1.south east) {feedback injected};

% Arrow 4: agent2 to feedback2
\draw[arrow] (agent2.south west) -- (feedback2.north east);
\node[font=\scriptsize\bfseries, text=gray, anchor=south east] at (feedback2.north east) {provisional tool call};

% Arrow 5: feedback2 to execute (feedback)
\draw[arrow, dashed, green!80] (feedback2.south east) -- (execute.north west);
\node[font=\scriptsize\bfseries, text=green!70!black, anchor=north east] at (feedback2.south east) {feedback injected};

\end{tikzpicture}
    }
    \caption{Example trajectory with inference-time feedback. The feedback agent (o3-mini) evaluates provisional tool calls from the tool-calling agent (GPT-4o) before execution. Loop 1: feedback provided. Loop 2: revised call approved.}
    \label{fig:trajectories}
\end{figure}

Large language models are increasingly deployed as agents that interact with external tools and APIs. These tool-calling agents face systematic challenges: selecting the correct tool, constructing calls with appropriate arguments, and recognizing when no tool can address a request \cite{patil2023gorillalargelanguagemodel,bfcl2024,kokane2025toolscanbenchmarkcharacterizingerrors}.

Two main classes of strategies address these challenges. \textbf{Training-based approaches} like GRPO \cite{tang2024generalizedpreferenceoptimizationunified} require substantial compute and slow deployment. \textbf{Inference-time approaches} like Self-Refine \cite{madaan2023selfrefineiterativerefinementselffeedback} and Reflexion \cite{shinn2023reflexionlanguageagentsverbal} enable self-correction without training, but require complex infrastructure and context management when agents must simultaneously generate and reflect on tool calls.

Both strategies face a fundamental \textbf{state recovery problem}. When an agent executes an incorrect action (such as deleting an alarm instead of updating it), self-correction requires maintaining the previous state in context. This becomes prohibitively expensive in complex execution environments and multi-turn scenarios, where the space of alternative trajectories grows exponentially. Without unreasonably large context windows (limited by model capacity and context budget), agents cannot reliably recover from destructive errors.

To address the native challenges of tool-calling agents and the state recovery problem, we propose \textbf{inference-time feedback} using a simple, configurable duo-agent architecture: a specialized reviewer agent evaluates provisional tool calls before execution and either provides feedback to the tool-calling agent or uses a selection strategy to choose among candidates. An example trajectory with the reviewer model is shown in Figure~\ref{fig:trajectories}. The key proposal is simple separation of concerns, which yields strong benefits. First, it requires only one additional agent (configurable by model and review strategy) rather than complex infrastructure changes. Second, by reviewing calls before execution, it helps mitigate destructive errors rather than attempting recovery, reducing the state recovery problem. The tool-calling agent requires no retraining or rearchitecture and seamlessly adopts feedback from the reviewer.

However, introducing a reviewer brings trade-offs: feedback can mitigate errors but can also break valid responses. To quantify this, we introduce \textbf{Helpfulness-Harmfulness metrics} quantifying how often feedback corrects errors versus introduces new ones. Reviewer quality can be improved through model capacity or prompt optimization. Latency overhead can be reduced through distillation.

To find the optimal feedback agent configuration, we explore multiple review strategies (progressive feedback, best-of-N selection, and best-of-N grading) and address reviewer failures through automated prompt optimization (APO). APO optimizes only the reviewer's prompt (the main agent's prompt remains unchanged), automatically refining it by observing cases where the reviewer made incorrect judgments.

For the reviewer agent, we compare non-reasoning (GPT-4o) and reasoning models to assess the impact of reasoning on review quality. We use o3-mini for initial experiments, then GPT-5 mini (adopted upon release) for APO experiments to leverage the latest reasoning capability. We chose mini variants to balance reasoning capability with cost efficiency. The main tool-calling agent remains GPT-4o throughout.

\paragraph{Key Results}
We evaluate on two benchmarks: BFCL (Berkeley Function-Calling Leaderboard) \cite{bfcl2024} for single-turn function calling and $\tau^2$-Bench \cite{barres2025tau2benchevaluatingconversationalagents} for multi-turn stateful scenarios. Our best configuration achieves +5.5\% on irrelevance detection (84.9\% $\rightarrow$ 90.4\%), +1.6\% on the relevance suite (90.9\% $\rightarrow$ 92.5\%) on BFCL (Table~\ref{tab:apo-results}), and +7.1\% on $\tau^2$-Bench (48.7\% $\rightarrow$ 55.8\%; Table~\ref{tab:tau2-complete}). Using our Helpfulness-Harmfulness metrics, we find that reasoning models (o3-mini) outperform standard language models as reviewers, achieving a 3:1 benefit-to-risk ratio (36.8\% helpfulness, 11.7\% harmfulness; Figure~\ref{fig:helpfulness-harmfulness}).

Reasoning model comparison and APO are evaluated on BFCL only; extending these to $\tau^2$-Bench remains future work.

This approach provides practical benefits for tool-calling systems: it requires no retraining or infrastructure modifications, supports rapid iteration on reviewer strategies through automated optimization, and offers tunable accuracy-latency trade-offs for different application requirements. The modular architecture enables organizations to enhance agent reliability incrementally while leaving existing tool-calling pipelines unchanged.

\paragraph{Key Contributions}
In summary, our work makes the following contributions:
\begin{enumerate}
\item \textbf{Inference-time feedback mechanism} improving tool-calling performance without training, achieving +5.5\% on irrelevance detection (BFCL) and +7.1\% on multi-turn scenarios ($\tau^2$-Bench).
\item \textbf{Helpfulness-Harmfulness metrics} quantifying benefit-risk tradeoffs of feedback interventions, showing reasoning models achieve 3:1 ratios versus standard language models (BFCL).
\item \textbf{Automated reviewer prompt optimization} systematically discovering effective review strategies via GEPA \cite{gepa2024}, achieving +1.5\% (relevance) and +2.8\% (irrelevance) on BFCL (Table~\ref{tab:apo-results}).
\end{enumerate}

% Method
\section{Method}

\begin{figure}[t]
\centering
\begin{tikzpicture}[
  node distance=0.6cm and 0.4cm,
  box/.style={rectangle, draw, rounded corners, minimum width=1.4cm, minimum height=0.5cm, align=center, font=\tiny, fill=white},
  arrow/.style={->, >=stealth, thick},
  label/.style={font=\tiny, align=center}
]
  % Tool-Calling Agent in center
  \node[box] (system) {Tool-Calling\\Agent};

  % Reviewer Agent bottom-left
  \node[box, below left=0.8cm and 0.5cm of system] (reinforce) {Reviewer\\Agent};

  % Execution Environment bottom-right
  \node[box, below right=0.8cm and 0.5cm of system] (exec) {Execution\\Environment};

  % Query/Response arrows (from/to external)
  \draw[arrow] ([yshift=0.6cm, xshift=-0.3cm]system.north) -- node[left, label] {Query} ([xshift=-0.3cm]system.north);
  \draw[arrow] ([xshift=0.3cm]system.north) -- node[right, label] {Response} ([yshift=0.6cm, xshift=0.3cm]system.north);

  % System <-> Feedback arrows (go to south west corner)
  \draw[arrow] ([xshift=-0.2cm]system.south west) -- node[left, label] {Provisional\\Tool Call} (reinforce.north);
  \draw[arrow] (reinforce.north east) -- node[right, label] {Feedback} ([xshift=0.1cm]system.south west);

  % System <-> Execution arrows (go to south east corner)
  \draw[arrow] ([xshift=0.2cm]system.south east) -- node[right, label] {Execute\\Tool Call} (exec.north);
  \draw[arrow] (exec.north west) -- node[left, label] {Result} ([xshift=-0.1cm]system.south east);

\end{tikzpicture}
\caption{Feedback Architecture. The reviewer agent reviews provisional tool calls before execution. If errors are detected, feedback is provided for revision. This loop continues until approval or maximum iterations (N) is reached.}
\label{fig:architecture}
\end{figure}
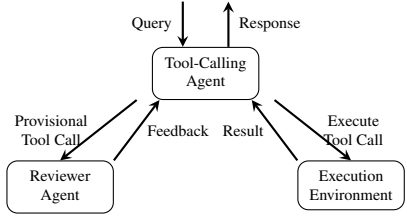

\subsection{Multi-Agent Architecture}
\label{sec:multi-agent-architecture}

We evaluate three collaboration mechanisms between a tool-calling agent and a reviewer agent:

\textbf{Progressive Feedback:} The feedback agent iteratively reviews the tool-calling agent's responses (Figure~\ref{fig:architecture}). If errors are found, feedback is injected as a system message and the tool-calling agent generates a revised response. This continues for up to N review loops or until no errors are detected. We denote this as \texttt{rN} (e.g., \texttt{r2} for up to 2 review loops).

\textbf{Best-of-N Selection (Selector):} The tool-calling agent generates N candidate responses with varying temperatures (0.3 to 1.0). The selector agent evaluates all candidates and selects the best one. This single-shot selection is denoted as \texttt{sN} (e.g., \texttt{s5} for 5 candidates).

\textbf{Best-of-N Grading (Grader):} Similar to selection, but the grader agent assigns explicit numeric scores (0.0-1.0) to each candidate with rationales. The highest-scored candidate is selected. Denoted as \texttt{gN} (e.g., \texttt{g5}).

We use a systematic naming convention for all mechanisms when reporting the evaluation results. For example, \texttt{4o-r2-5-mini-v3-gepa} indicates GPT-4o base model, progressive feedback with up to 2 feedback loops (r2), GPT-5 mini feedback model (5-mini), GEPA prompt version 3. See Appendix~\ref{sec:appendix-mechanism-examples} for concrete examples of how each mechanism operates.

\subsection{Reviewer Prompt Optimization}
\label{sec:apo}

Manual prompt engineering for reviewer agents is labor-intensive and may miss subtle failure patterns. Inspired by GEPA \cite{gepa2024}, which iteratively improves prompts based on execution feedback, we automatically refine reviewer prompts by observing cases where the reviewer made incorrect judgments. Starting from manually-refined v2 prompts for BFCL, we iteratively improve them using a reasoning model for reflection. Applying this optimization to $\tau^2$-Bench prompts remains future work. Details and results appear in Section~\ref{sec:experiments-apo}.

% Experiment Setup
\section{Experiment Setup}

\subsection{Benchmarks}

We evaluate on two benchmarks: BFCL (Berkeley Function Calling Leaderboard) for single-turn tool calling and $\tau^2$-Bench for multi-turn stateful scenarios.

\subsubsection{BFCL}

Single-turn, stateless tool calling. We evaluate on Non-Live (BFCL V1, curated) and Live (BFCL V2, community-contributed) categories~\cite{bfcl2024}. Categories include \texttt{simple}, \texttt{multiple}, \texttt{parallel}, and \texttt{parallel\_multiple} (combined parallel and sequential; hardest), which together form the \texttt{relevance suite}. The \texttt{irrelevance} category tests detecting when no tool is relevant.

\subsubsection{\texorpdfstring{$\tau^2$-Bench}{Tau2-Bench}}

Multi-turn, stateful tool calling with domain-specific policies across three domains (airline, retail, telecom). Agents must maintain conversational context, verify state preconditions, and handle benchmark-specific constraints.

\subsection{Models}

All experiments use GPT-4o (gpt-4o-2024-11-20 snapshot) as the base tool-calling agent with temperature=0 and seed=42 for reproducibility. Initial experiments use o3-mini as the reasoning model reviewer; APO experiments use GPT-5 mini (adopted upon release to leverage the latest reasoning capability). Both reasoning models use reasoning\_effort=medium.

% Results & Analysis
\section{Results \& Analysis}

We organize our evaluation around three research questions:
\begin{itemize}
\item \textbf{RQ1 (Effectiveness \& Error):} What is the effectiveness of inference-time feedback for tool-calling agents, and what are the associated error correction trade-offs?
\item \textbf{RQ2 (Design \& Optimization):} How do feedback mechanism design, reviewer model selection, and automated optimization affect reviewer agent performance?
\item \textbf{RQ3 (Latency \& Deployment):} What are the latency overhead and deployment trade-offs of inference-time feedback across different application scenarios?
\end{itemize}
We answer these questions using both standard benchmark metrics and additional metrics that quantify error correction versus error introduction (Section~\ref{sec:eval-metrics}). We then address RQ1 by evaluating effectiveness on BFCL and $\tau^2$-Bench (Section~\ref{sec:effectiveness}), RQ2 by comparing reviewer models, feedback mechanisms, and automated prompt optimization (Section~\ref{sec:design-optimization}), and RQ3 by analyzing latency overhead and deployment trade-offs (Section~\ref{sec:latency-deployment}).

\subsection{Evaluation Metrics}
\label{sec:eval-metrics}

We evaluate using each benchmark's default metrics: per-category accuracy for BFCL (simple, multiple, parallel, parallel\_multiple, irrelevance) and relevance suite (unweighted average of the first four), and per-domain pass rate for $\tau^2$-Bench (airline, retail, telecom). To complement these standard metrics, we introduce three metrics that quantify reviewer quality and error correction trade-offs:
\begin{itemize}
\item \textbf{Helpfulness:} Percentage of test cases where base agent is wrong AND reviewer agent corrects it.
\item \textbf{Harmfulness:} Percentage of test cases where base agent is right AND reviewer introduces error.
\item \textbf{Benefit-to-Risk Ratio:} Helpfulness ÷ Harmfulness.
\end{itemize}
These metrics reveal whether feedback provides net positive value, showing not just final accuracy but how the reviewer affects correct and incorrect responses.

\subsection{Effectiveness \& Error Trade-offs}
\label{sec:effectiveness}

\textit{RQ1: What is the effectiveness of inference-time feedback for tool-calling agents, and what are the associated error correction trade-offs?}

\subsubsection{BFCL Evaluation}

We first evaluate inference-time feedback on BFCL to establish baseline effectiveness on single-turn, stateless tool calling. Results are averaged across experimental sessions.

\paragraph{Initial Results}

With minimal v1 prompts and GPT-4o as reviewer, we improved irrelevance detection but yielded only marginal gains on other categories.

\begin{table}[ht]
\centering
\small
\begin{tabular}{lrrr}
\toprule
\textbf{Category} & \textbf{Baseline} & \textbf{+ Reviewer} & \textbf{$\Delta$} \\
\midrule
Simple & 92.4\% & 92.8\% & +0.4\% \\
Multiple & 92.8\% & 93.0\% & +0.2\% \\
\textbf{Irrelevance} & \textbf{84.9\%} & \textbf{89.6\%} & \textbf{+4.7\%} \\
Rel. Suite & 90.9\% & 91.4\% & +0.5\% \\
\bottomrule
\end{tabular}
\caption{Initial Reviewer Impact (4o-r5-4o-v1).}
\end{table}

\begin{figure}[ht]
\centering
\resizebox{0.98\columnwidth}{!}{\begin{tikzpicture}[
    node distance=0.4cm,
    block/.style={rectangle, rounded corners, draw=black, thick, text width=5.8cm, minimum height=0.8cm, align=left, font=\sffamily\scriptsize},
    agent/.style={block, fill=blue!15, draw=blue!80},
    reviewer/.style={block, fill=orange!15, draw=orange!80},
    label/.style={font=\sffamily\scriptsize\bfseries, text width=1.2cm, align=right},
    arrow/.style={-Stealth, thick, gray}
]

% Tool Agent Block
\node (AgentBlock) [agent] {
    \texttt{\{"tool\_calls": [\{"name": "get\_weather",}\\
    \texttt{\ \ "arguments": \{"city": "Seattle"\}\}]\}}
};
\node[label, left=0.2cm of AgentBlock, text=blue!80] {Tool\\Agent};

% Arrow
\draw[arrow] (AgentBlock.south) -- ++(0,-0.4cm) node[midway, right, font=\scriptsize\itshape, text=gray] {Sent for review};

% Reviewer Block
\node (ReviewerBlock) [reviewer, below=0.6cm of AgentBlock] {
    ``Response lacks follow-up explanation for the user about what the weather information means.''
};
\node[label, left=0.2cm of ReviewerBlock, text=orange!80] {Reviewer\\Agent};

\end{tikzpicture}}
\caption{Reviewer misunderstanding example. The reviewer expects user-facing dialogue, but BFCL only evaluates tool call accuracy.}
\label{fig:reviewer-misunderstanding}
\end{figure}
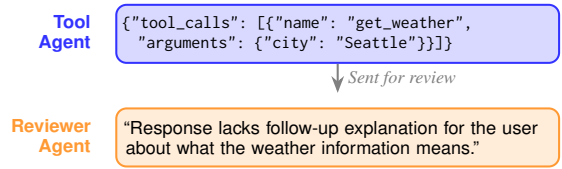

\paragraph{Root Cause Analysis} The reviewer incorrectly flagged valid tool calls as ``incomplete,'' expecting execution results. However, as shown in Figure~\ref{fig:architecture}, the reviewer evaluates provisional tool calls before execution, so no results exist yet. Analysis showed 23\% of cases had redundant iterations where the reviewer demanded elaboration it should not expect. Figure~\ref{fig:reviewer-misunderstanding} illustrates a typical example of this mismatch.

\paragraph{Addressing Over-Skepticism}
\label{sec:over-skepticism}

We added an explicit guideline addressing the over-skepticism failure mode:

\begin{quote}
\small
\raggedright
\textbf{[CRITICAL] Tool-only responses are complete.} Do not mark tool-only responses as incomplete for lacking user-facing answers, follow-up explanations, or final results. Tool calls are standalone steps. Focus on whether the actual tool calls are correct.
\end{quote}

The explicit guidance reduced redundant review loops from 23\% to 8\%, improving efficiency. However, v2 prompts reduced harmfulness but did not fully solve the irrelevance detection challenge. This motivated comparing different review models.

\begin{table}[ht]
\centering
\small
\resizebox{\columnwidth}{!}{%
\begin{tabular}{llrrrrr}
\toprule
\textbf{Reviewer} & \textbf{Prompt} & \textbf{Rel.} & \textbf{Irrel.} & \textbf{Help.} & \textbf{Harm.} & \textbf{Ratio} \\
\midrule
GPT-4o & v1 & 89.5\% & 90.0\% & 30.2\% & 14.2\% & 2.1:1 \\
GPT-4o & v2 & 91.0\% & 90.4\% & 34.9\% & 12.9\% & 2.7:1 \\
o3-mini & v2 & \textbf{91.8\%} & \textbf{91.0\%} & \textbf{36.8\%} & \textbf{11.7\%} & \textbf{3.1:1} \\
\bottomrule
\end{tabular}%
}
\caption{Helpfulness vs. Harmfulness on BFCL Non-Live. Results from dedicated experiment measuring reviewer quality (see Figure~\ref{fig:helpfulness-harmfulness}). Rel. = relevance suite, Irrel. = irrelevance, Help. = helpfulness, Harm. = harmfulness.}
\label{tab:feedback-comparison}
\end{table}

\begin{figure}[b]
\centering
\includegraphics[width=0.85\linewidth]{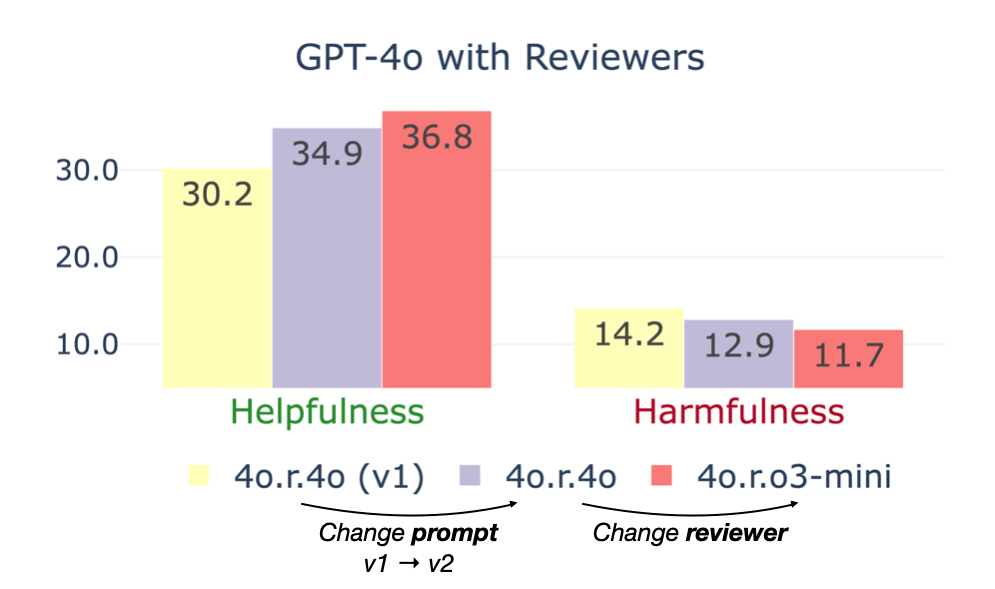}
\caption{Helpfulness vs. Harmfulness trade-off on BFCL Non-Live. o3-mini achieves the best balance (3.1:1 ratio).}
\label{fig:helpfulness-harmfulness}
\end{figure}

Table~\ref{tab:feedback-comparison} compares o3-mini and GPT-4o as reviewer models with manually-engineered prompts (v1, v2; see Appendix~\ref{sec:appendix-prompts} for full prompts). The o3-mini configuration achieves the best performance: 91.8\% on the relevance suite and 91.0\% on irrelevance detection. Beyond accuracy, the helpfulness and harmfulness metrics reveal critical differences in reviewer quality. The o3-mini configuration corrects 36.8\% of base agent errors while introducing errors in only 11.7\% of previously correct responses, achieving a 3.1:1 benefit-to-risk ratio (Figure~\ref{fig:helpfulness-harmfulness}). A detailed analysis of model comparison is presented in Section~5.2.

\subsection{\texorpdfstring{$\tau^2$-Bench}{Tau2-Bench} Evaluation}

We extend evaluation to $\tau^2$-Bench to test whether feedback mechanisms generalize to multi-turn, stateful scenarios with domain-specific policies. Results are averaged over 3 benchmark runs.

\paragraph{Initial Results}

Table~\ref{tab:tau2-results} shows performance across three domains. The best configuration (4o-r5-4o-v1) achieves +7.1\% average improvement over baseline (48.7\% → 55.8\%).

\begin{table}[ht]
\centering
\small
\resizebox{\columnwidth}{!}{%
\begin{tabular}{lrrrr}
\toprule
\textbf{Configuration} & \textbf{Airline} & \textbf{Retail} & \textbf{Telecom} & \textbf{Average} \\
\midrule
4o baseline & 42.0\% & 62.9\% & 41.2\% & 48.7\% \\
\midrule
\multicolumn{5}{l}{\textit{Progressive Feedback (rN)}} \\
4o-r5-4o-v1 & 40.7\% & 62.6\% & \textbf{64.0\%} & \textbf{55.8\%} \\
4o-r5-4o-v2 & 40.7\% & 58.5\% & 59.6\% & 52.9\% \\
\midrule
\multicolumn{5}{l}{\textit{Best-of-N Selection (sN)}} \\
4o-s5-4o-v1 & 42.7\% & 61.1\% & 38.0\% & 47.3\% \\
4o-s5-4o-v2 & 48.0\% & 61.4\% & 48.2\% & 52.5\% \\
\midrule
\multicolumn{5}{l}{\textit{Best-of-N Grading (gN)}} \\
4o-g5-4o-v1 & 47.3\% & 60.5\% & 45.3\% & 51.0\% \\
4o-g5-4o-v2 & 46.7\% & \textbf{65.8\%} & 48.8\% & 53.8\% \\
\bottomrule
\end{tabular}
}
\caption{Performance on $\tau^2$-Bench across all mechanisms. Progressive Feedback achieves the highest average (55.8\%). See Table~\ref{tab:tau2-complete} for additional details.}
\label{tab:tau2-results}
\end{table}

\paragraph{Error analysis} 
We analyzed failures to understand where feedback helps and hurts. Table~\ref{tab:tau2-errors} shows distribution.

\begin{table}[ht]
\centering
\small
\resizebox{\columnwidth}{!}{%
\begin{tabular}{lrr}
\toprule
\textbf{Error Type} & \textbf{Baseline} & \textbf{With Reviewer} \\
\midrule
Policy Constraint Violation & 31\% & 18\% (-13\%) \\
Missing Context Awareness & 24\% & 15\% (-9\%) \\
Incorrect Tool Selection & 19\% & 22\% (+3\%) \\
Argument Errors & 16\% & 18\% (+2\%) \\
Over-verbalization & 10\% & 27\% (+17\%) \\
\bottomrule
\end{tabular}
}
\caption{Failure Mode Distribution on $\tau^2$-Bench}
\label{tab:tau2-errors}
\end{table}

For example, in an airline booking domain, the policy requires checking seat availability before booking. Base agent error: directly calls \texttt{book\_flight()} without calling \texttt{check\_availability()} first. Reviewer agent catches: ``Response violates state precondition: seat availability must be checked.''

The reviewer effectively catches policy violations (-13\%) but introduces new ``over-verbalization'' errors (+17\%). The over-skepticism problem observed in BFCL recurs here: the reviewer flags tool-only responses in contexts where $\tau^2$-Bench expects mixed responses. Comparing mechanisms, Progressive Feedback substantially outperforms Best-of-N approaches (+3--8\% on average), with selection actually underperforming the baseline on some domains (Table~\ref{tab:tau2-results}).

\textbf{Cross-benchmark generalization:} Applying BFCL-optimized prompts (v2-bfcl) directly to $\tau^2$-Bench introduces errors, highlighting the domain mismatch between single-turn function calling and multi-turn stateful agents. The different evaluation criteria and interaction patterns require benchmark-specific prompt adaptation.

\paragraph{Addressing Domain Constraints}

$\tau^2$-specific prompts (v2-tau; see Appendix~\ref{sec:appendix-prompts}) emphasize context awareness (marked CRITICAL), state preconditions, and benchmark-specific constraints. However, baseline reviewer prompts (v1) sometimes outperform domain-tuned reviewer prompts on average across domains (55.8\% vs. 52.9\%; Table~\ref{tab:tau2-results}), suggesting that manually-engineered instructions may not generalize across all tasks within a domain. This opens an opportunity for automated prompt optimization on $\tau^2$-Bench as future work.

\subsection{Design \& Optimization}
\label{sec:design-optimization}

\textit{RQ2: How do feedback mechanism design, reviewer model selection, and automated optimization affect reviewer agent performance?}

\subsubsection{Model Comparison}

On the BFCL benchmark, we compare o3-mini and GPT-4o as reviewer models with manually-engineered prompts (v1, v2; see Appendix~\ref{sec:appendix-prompts} for full prompts) for initial experiments, then GPT-5 mini (its successor, as the more up-to-date reasoning model) for automated reviewer prompt optimization using GEPA (v3-gepa).

The reasoning model's advantage manifests in two key areas. First, o3-mini outperforms GPT-4o on irrelevance detection by +0.6\% (91.0\% vs. 90.4\%), a category that requires determining whether any available tool can address the request. Second, o3-mini exhibits lower harmfulness (11.7\% vs. 12.9\%; Table~\ref{tab:feedback-comparison}), making fewer false positive errors. The reasoning model's systematic verification reduces over-correction, leading to more reliable feedback.

\textbf{Irrelevance detection:} o3-mini outperforms GPT-4o by +0.6\% on irrelevance (91.0\% vs. 90.4\% for v2 prompts; Table~\ref{tab:feedback-comparison}). This category requires determining whether any available tool can address the request.

\textbf{Error introduction rates:} o3-mini exhibits lower harmfulness (11.7\% vs. 12.9\% for GPT-4o v2; Table~\ref{tab:feedback-comparison}), making fewer false positive errors. The reasoning model's systematic verification reduces over-correction.

\textbf{Key insight:} The goal is maximizing helpfulness while minimizing harmfulness (Figure~\ref{fig:helpfulness-harmfulness}). The o3-mini configuration achieves the best balance: 36.8\% of base agent errors corrected with 11.7\% new errors introduced, achieving a 3.1:1 benefit-to-risk ratio. This demonstrates that reasoning models provide better feedback quality than standard language models.

\subsubsection{Mechanism Comparison}

We evaluate three feedback mechanisms on BFCL to understand their strengths: Progressive Feedback (iterative refinement with feedback), Best-of-N Selection (choosing the best response from multiple candidates), and Best-of-N Grading (scoring and selecting from multiple candidates). Appendix Table~\ref{tab:bfcl-complete} provides complete results across all mechanisms and categories.

\textbf{Reasoning model advantage:} As shown in Table~\ref{tab:feedback-comparison}, o3-mini achieves the best helpfulness-to-harmfulness ratio (3.1:1) compared to GPT-4o (2.7:1 for v2 prompts). This suggests that reasoning models provide more reliable feedback with fewer false corrections.

\textbf{Mechanism effectiveness:} Progressive Feedback outperforms Best-of-N approaches on irrelevance detection by +4--5\%, while Best-of-N shows only marginal gains on relevance. Progressive Feedback's advantage lies in explicitly identifying and correcting errors through targeted feedback, particularly for irrelevance where the model must determine no tool can address the request.

\subsubsection{Automatic Context Optimization}
\label{sec:experiments-apo}

Having established effectiveness on BFCL and demonstrated generalization on $\tau^2$-Bench, we now systematize improvements through automated reviewer prompt optimization. We apply GEPA to BFCL reviewer prompts; extending this to $\tau^2$-Bench remains future work.

\paragraph{GEPA Algorithm and Results}

GEPA (Genetic-Pareto prompt evolution) addresses limitations of manual prompt engineering. The algorithm starts with manually-engineered reviewer prompts (v2), collects failure cases, uses an LLM to reflect and propose improvements, and iterates until convergence. This produces v3 prompts approximately 4.5$\times$ longer (1,599 vs.\ 358 tokens)\footnote{Measured using OpenAI's tokenizer (\url{https://platform.openai.com/tokenizer}) with the GPT-5.x \& O1/3 option.} with detailed error criteria, edge case handling, and error checklists (Appendix~\ref{sec:appendix-prompts}).

\begin{table}[ht]
\centering
\small
\resizebox{\columnwidth}{!}{%
\begin{tabular}{lrrr}
\toprule
\textbf{Configuration} & \textbf{Rel.} & \textbf{Irrel.} & \textbf{Gain} \\
\midrule
4o-r2-5-mini-v2 & 91.0\% & 87.6\% & baseline \\
4o-r2-5-mini-v3-gepa & 92.5\% & 90.4\% & +1.5\% / +2.8\% \\
\bottomrule
\end{tabular}
}
\caption{GEPA Optimization on BFCL Non-Live (Progressive Feedback). v2 = manual, v3-gepa = GEPA-optimized. Rel. = relevance suite, Irrel. = irrelevance.}
\label{tab:apo-results}
\end{table}

The GEPA-optimized prompts show improvements across categories, with particularly strong gains on the parallel\_multiple category (+2.1\%; Table~\ref{tab:bfcl-complete}). This category involves orchestrating multiple parallel tool calls with correct argument passing, where APO-optimized feedback catches mismatches and missing calls, demonstrating automated optimization discovers strategies difficult to anticipate manually.

\subsection{Latency \& Deployment Trade-offs}
\label{sec:latency-deployment}

\textit{RQ3: What are the latency overhead and deployment trade-offs of inference-time feedback across application scenarios?}

Inference-time feedback introduces computational overhead that varies by task type. We analyze latency measurements from our experiments to characterize the cost-accuracy trade-off.

\paragraph{Latency Analysis:} 

\begin{figure}[t]
\centering
\includegraphics[width=\columnwidth]{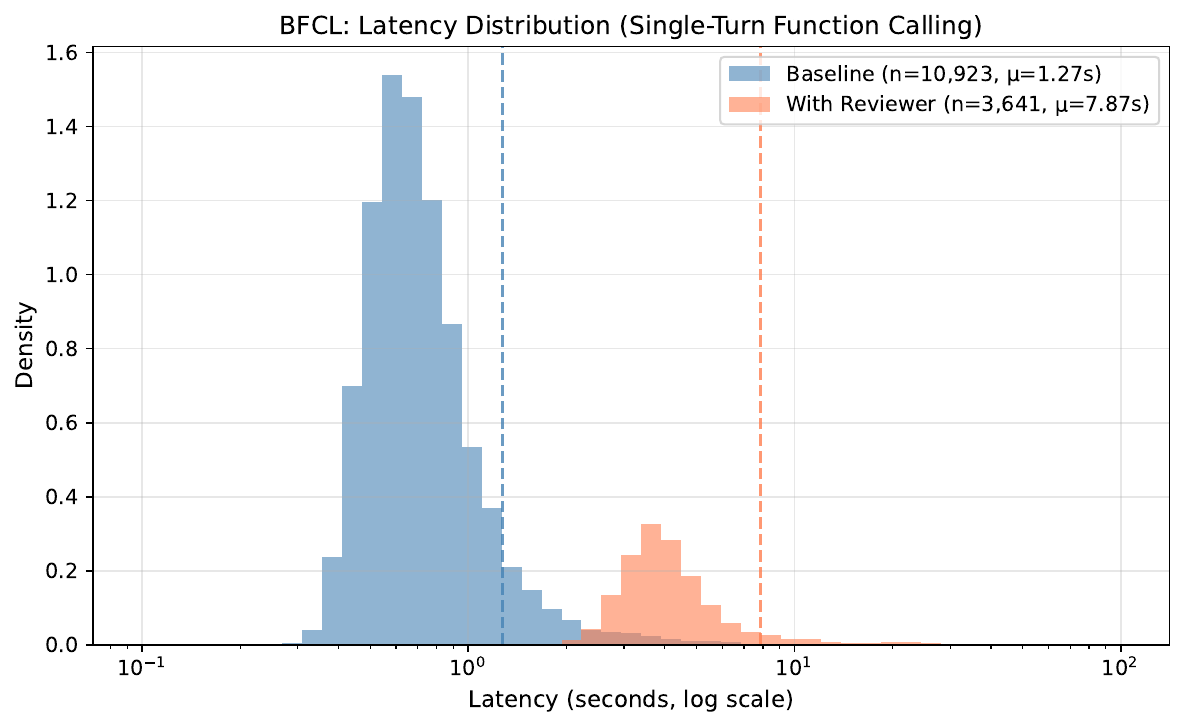}
\includegraphics[width=\columnwidth]{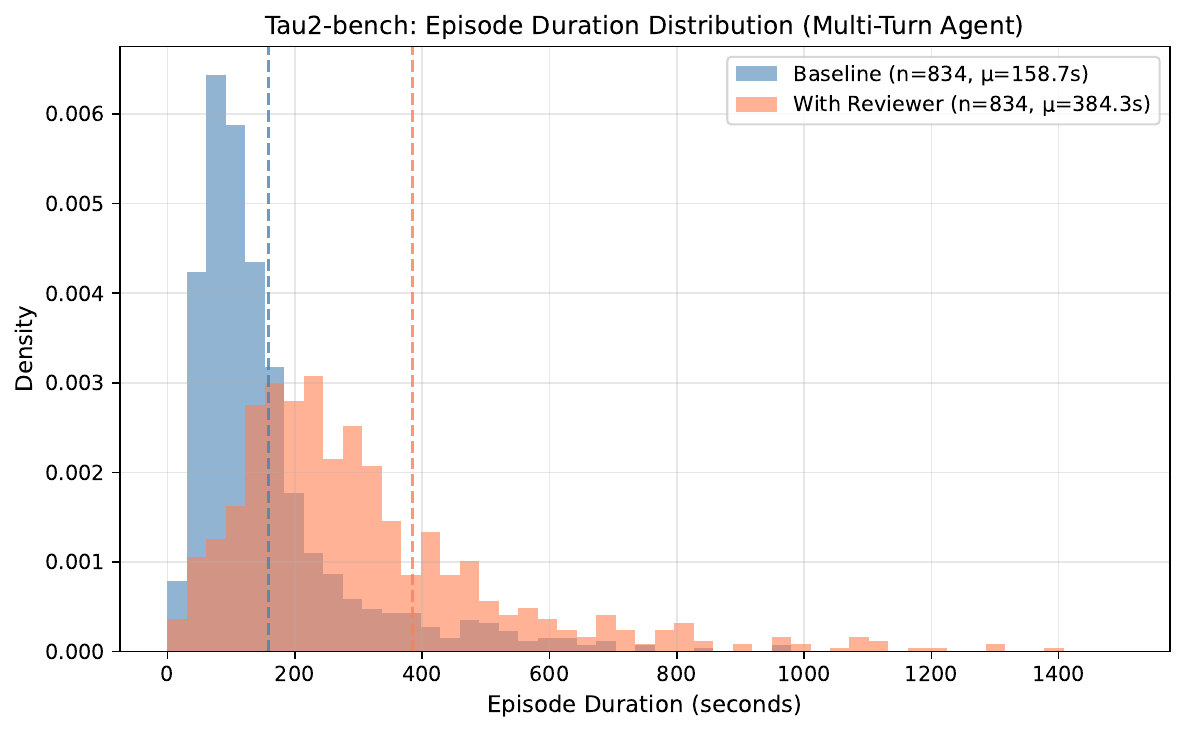}
\caption{Latency distributions. Top: BFCL per-item latency (log scale). Bottom: $\tau^2$-Bench per-episode duration. Blue = baseline, coral = with reviewer. Dashed lines = means.}
\label{fig:latency-main}
\end{figure}

We measure latency impact on both BFCL (single-turn function calling) and $\tau^2$-Bench (multi-turn agent episodes). On BFCL, the feedback mechanism (r5-4o) increases average latency from 1.27s to 7.87s, a \textbf{6.2× multiplier}. This increase occurs because the baseline is a single inference call, so reviewer overhead dominates. On $\tau^2$-Bench, the same mechanism increases average episode duration from 158.7s to 384.3s, a \textbf{2.4× multiplier}. The lower relative impact reflects the multi-turn nature: reviewer overhead is amortized across approximately 40 turns per episode. Figure~\ref{fig:latency-main} shows the latency distributions.

\paragraph{Reviewer Call Patterns:} On BFCL, feedback averages 1.33 reviewer calls per item. On $\tau^2$-Bench, the mechanism averages 0.96 reviewer calls per turn. The per-turn call count is lower for $\tau^2$-Bench likely due to stateful nature reducing ambiguity in later turns.

\paragraph{Application-Specific Deployment:} The latency-accuracy trade-off allows flexible deployment strategies:
\begin{itemize}
    \item \textbf{Single-turn, high-volume applications:} The 6.2× latency overhead may be prohibitive for real-time use cases. Consider baseline deployment or selective feedback on uncertain cases.
    \item \textbf{Multi-turn agents:} The 2.4× overhead is more acceptable when amortized across conversation turns. Feedback mechanisms are viable for complex, accuracy-critical workflows.
    \item \textbf{API-intensive workflows:} Feedback may achieve ROI-positive gains by preventing spurious API calls that incur downstream costs.
\end{itemize}

% % OLD SECTIONS - TO BE REMOVED LATER
% % Experiments (marked for removal)
% \input{experiments}

% % Discussion (marked for removal)
% \input{discussion}
% % TODO: Remove this section after migrating content

% Related Work
\section{Related Work}

\textbf{Tool-Calling Benchmarks.} BFCL \cite{bfcl2024,patil2023gorillalargelanguagemodel} evaluates single-turn function calling across categories. ToolSandbox \cite{lu2025toolsandboxstatefulconversationalinteractive} introduces stateful, conversational evaluation. $\tau^2$-Bench tests multi-turn tool calling with domain-specific policies. Our work is the first to systematically compare feedback mechanisms across both stateless (BFCL) and stateful ($\tau^2$-Bench) benchmarks.

\textbf{Agent Feedback and Self-Refinement.} Self-Refine \cite{madaan2023selfrefineiterativerefinementselffeedback} and Reflexion \cite{shinn2023reflexionlanguageagentsverbal} use self-feedback for iterative refinement. Our work differs by using a specialized reviewer agent instead of self-feedback, demonstrating that external feedback from reasoning models outperforms self-correction.

\textbf{Prompt Optimization.} MIPROv2 \cite{opsahlong2024optimizinginstructionsdemonstrationsmultistage} uses Bayesian optimization for instructions and few-shot examples. GEPA \cite{gepa2024} uses genetic-Pareto evolution with LLM reflection. We apply GEPA to reviewer agent prompts, showing improvements over manual engineering.

\textbf{Training-Based Approaches.} GRPO \cite{tang2024generalizedpreferenceoptimizationunified} fine-tunes via weight-space RL but requires thousands of rollouts. Recent reasoning models (o1, o3) \cite{openai2024o1} demonstrate strong verification capabilities. Knowledge distillation \cite{hinton2015distillingknowledgeneuralnetwork,ho2023largelanguagemodelsreasoning,fu2023specializingsmallerlanguagemodels} provides a foundation for future work on distilling reviewer agents. Our inference-time approach provides immediate gains without training.

% Conclusion
\section{Conclusion}

We introduce Reinforced Agent, an inference-time feedback mechanism for tool-calling agents. Evaluation on BFCL establishes baseline effectiveness and identifies over-skepticism as the primary reviewer error mode. Evaluation on $\tau^2$-Bench demonstrates generalization to multi-turn scenarios. Our Helpfulness-Harmfulness metrics show reasoning models achieve favorable benefit-to-risk ratios as reviewers. Automated prompt optimization via GEPA systematizes reviewer improvement over manual engineering. Latency overhead is most viable for complex, accuracy-critical workflows where it amortizes across turns, and can be further mitigated by distilling the reviewer into a smaller, faster model (e.g., a lightweight reward model or classifier) suitable for local or on-device deployment. This work establishes a practical path from benchmarking to optimization to deployment, with potential for distillation into reward models for reinforcement learning.

\section{Limitations}

Our work has several limitations.

\textbf{Base model scope.} We evaluate only GPT-4o as the base tool-calling agent and compare two reviewer models (GPT-4o and o3-mini). While the modular architecture imposes no constraints on the base model (the reviewer operates on tool-call outputs regardless of which model generates them), generalization to open-source models (e.g., Llama, Mistral) and smaller proprietary models remains empirically unvalidated.

\textbf{Analysis scope.} GEPA-based prompt optimization and Helpfulness-Harmfulness metrics were applied only to BFCL; extending these to multi-turn benchmarks like $\tau^2$-Bench requires adapting for partial-credit scoring and multi-turn error propagation. Additionally, generic prompts (v1) sometimes outperform domain-specific prompts (v2-tau) on $\tau^2$-Bench, suggesting manual tuning may not generalize without automated optimization.

\section{Ethics Statement}

To the best of our knowledge, all results published in this paper are accurate. All data sources are publicly available benchmarks and are cited accordingly. No human subjects, private user data, or personally identifiable information were used in this work.

\section*{Acknowledgments}

We thank Jaechan Lee for contributions to the exploratory experiments in this study during his internship at Apple. We are grateful to Sylvia Xu and Nishant Kanakia for detailed feedback on multiple drafts of this paper. We also thank Anatoly Adamov, Alex Braunstein, and Amar Subramanya for their continued support of this research.

% Acknowledgments
% \input{acknowledgments}

\bibliography{references}

% Appendix
\appendix

\section{Complete Experimental Results}
\label{sec:appendix}

\subsection{BFCL Results}
\label{sec:appendix-bfcl}

Table~\ref{tab:bfcl-complete} presents complete results on the Berkeley Function Calling Leaderboard (BFCL) across all configurations tested. We report both Non-Live (BFCL V1, curated) and Live (BFCL V2, community-contributed) categories.\footnote{BFCL V1 contains curated benchmark tasks; V2 adds community-contributed real-world data. See \url{https://huggingface.co/datasets/gorilla-llm/Berkeley-Function-Calling-Leaderboard} for dataset details.}

\begin{table*}[ht]
\centering
\small
\begin{tabular}{lrrrrrr}
\toprule
\textbf{Configuration} & \textbf{Simple} & \textbf{Multiple} & \textbf{Parallel} & \textbf{Par\_Mult} & \textbf{Irrel.} & \textbf{Rel. Suite} \\
\midrule
4o (baseline) & 92.4 & 92.8 & 93.0 & 85.2 & 84.9 & 90.9 \\
\midrule
\multicolumn{7}{l}{\textit{Progressive Feedback (Non-Live)}} \\
4o-r5-4o-v1 & 92.8 & 93.0 & 92.5 & 87.5 & 89.6 & 91.4 \\
4o-r5-4o-v2-bfcl & 93.0 & 94.5 & 92.5 & 86.0 & 84.6 & 91.5 \\
\midrule
\multicolumn{7}{l}{\textit{Best-of-N Selection (Selector)}} \\
4o-s5-4o-v1 & 92.5 & 93.5 & 92.5 & 87.0 & 85.8 & 91.4 \\
4o-s5-4o-v2-bfcl & 93.2 & 93.5 & 93.0 & 87.5 & 85.4 & 91.8 \\
\midrule
\multicolumn{7}{l}{\textit{Best-of-N Grading (Grader)}} \\
4o-g5-4o-v1 & 92.8 & 93.5 & 91.5 & 87.5 & 85.8 & 91.3 \\
4o-g5-4o-v2-bfcl & 94.0 & 92.5 & 92.5 & 86.5 & 85.8 & 91.4 \\
\midrule
\multicolumn{7}{l}{\textit{Automated Prompt Optimization (Non-Live)}} \\
4o-r2-5-mini-v1-1 & 92.2 & 93.2 & 91.7 & 85.2 & 88.8 & 90.5 \\
4o-r2-5-mini-v2-bfcl & 92.4 & 93.8 & 92.2 & 85.5 & 87.6 & 91.0 \\
4o-r2-5-mini-v3-gepa & \textbf{95.3} & \textbf{94.3} & \textbf{93.0} & \textbf{87.3} & \textbf{90.4} & \textbf{92.5} \\
\bottomrule
\end{tabular}
\caption{Complete BFCL Non-Live Results. All scores are percentages. \textit{APO}: evaluates GEPA-optimized prompts using GPT-5 mini. Agent naming: \texttt{\{base\}-\{mechanism\}\{N\}-\{feedback\_model\}-\{prompt\_version\}}.}
\label{tab:bfcl-complete}
\end{table*}

\begin{table}[ht]
\centering
\small
\resizebox{\columnwidth}{!}{%
\begin{tabular}{lrrrr}
\toprule
\textbf{Configuration} & \textbf{Simple} & \textbf{Mult.} & \textbf{Irrel.} & \textbf{Rel.} \\
\midrule
4o (baseline) & 79.8 & 78.9 & 77.6 & 79.2 \\
\midrule
\multicolumn{5}{l}{\textit{o3-mini Reviewer}} \\
4o-r-o3-mini & 82.2 & 79.0 & 80.8 & 79.6 \\
\midrule
\multicolumn{5}{l}{\textit{Automated Prompt Optimization (APO)}} \\
4o-r2-5-mini-v1-1 & 80.6 & 77.6 & 83.2 & 78.1 \\
4o-r2-5-mini-v2-bfcl & 80.9 & 78.4 & 80.8 & 78.9 \\
4o-r2-5-mini-v3-gepa & \textbf{83.1} & 77.4 & \textbf{83.6} & 78.5 \\
\midrule
\multicolumn{5}{l}{\textit{Progressive Feedback (GPT-4o feedback)}} \\
4o-r5-4o-v1 & 78.7 & 76.8 & 82.2 & 77.5 \\
4o-r5-4o-v2-bfcl & 80.6 & 78.8 & 76.4 & 79.3 \\
\midrule
\multicolumn{5}{l}{\textit{Best-of-N Selection (Selector)}} \\
4o-s5-4o-v1 & 81.0 & 79.1 & 79.5 & 79.6 \\
4o-s5-4o-v2-bfcl & 81.8 & \textbf{79.2} & 78.8 & \textbf{79.8} \\
\midrule
\multicolumn{5}{l}{\textit{Best-of-N Grading (Grader)}} \\
4o-g5-4o-v1 & 81.4 & 78.7 & 79.5 & 79.5 \\
4o-g5-4o-v2-bfcl & 79.8 & 78.9 & 78.7 & 79.2 \\
\bottomrule
\end{tabular}
}
\caption{BFCL Live Results. Live categories are community-contributed and generally more challenging than Non-Live. Parallel categories omitted due to limited Live data.}
\label{tab:bfcl-live}
\end{table}

\subsection{\texorpdfstring{$\tau^2$-Bench}{Tau2-Bench} Results}
\label{sec:appendix-tau2}

Table~\ref{tab:tau2-complete} presents complete results on $\tau^2$-Bench across all configurations tested.

\begin{table}[ht]
\centering
\small
\begin{tabular}{lrrrr}
\toprule
\textbf{Configuration} & \textbf{Airline} & \textbf{Retail} & \textbf{Telecom} & \textbf{Avg} \\
\midrule
4o (baseline) & 42.0 & 62.9 & 41.2 & 48.7 \\
\midrule
\multicolumn{5}{l}{\textit{Progressive Feedback}} \\
4o-r5-4o-v1 & 40.7 & 62.6 & \textbf{64.0} & \textbf{55.8} \\
4o-r5-4o-v2-tau & 40.7 & 58.5 & 59.6 & 52.9 \\
\midrule
\multicolumn{5}{l}{\textit{Best-of-N Selection}} \\
4o-s5-4o-v1 & 42.7 & 61.1 & 38.0 & 47.3 \\
4o-s5-4o-v2-tau & 48.0 & 61.4 & 48.2 & 52.5 \\
\midrule
\multicolumn{5}{l}{\textit{Best-of-N Grading}} \\
4o-g5-4o-v1 & 47.3 & 60.5 & 45.3 & 51.0 \\
4o-g5-4o-v2-tau & \textbf{46.7} & \textbf{65.8} & 48.8 & 53.8 \\
\bottomrule
\end{tabular}
\caption{Complete $\tau^2$-Bench Results. All scores are percentages. Best results per column in bold.}
\label{tab:tau2-complete}
\end{table}

\subsection{Feedback Mechanism Examples}
\label{sec:appendix-mechanism-examples}

This section illustrates how each feedback mechanism operates with concrete examples using a weather query task.

\subsubsection{Progressive Feedback (r2): Iterative Refinement}

Progressive feedback allows up to 2 review loops. The process continues until the feedback agent accepts the response or the maximum loops are reached.

\textbf{Example execution:}

\textit{User Query:} ``What's the weather in New York City?''

\textbf{Loop 1:}
\begin{itemize}
    \item \textit{GPT-4o generates:} \texttt{get\_weather(location="NYC", temp\_unit="celsius")}
    \item \textit{o3-mini reviews:} ``Error: For US cities, temperature should use Fahrenheit by default.''
\end{itemize}

\textbf{Loop 2:}
\begin{itemize}
    \item \textit{GPT-4o revises with feedback:} \texttt{get\_weather(location="New York", temp\_unit="fahrenheit")}
    \item \textit{o3-mini reviews:} ``Correct. Tool call is properly formatted with appropriate units.''
    \item \textit{Result:} Accepted. Process terminates after 2 loops.
\end{itemize}

\subsubsection{Best-of-N Grading (g5): Scored Selection}

The grader mechanism generates 5 candidate responses at different temperatures and assigns explicit scores to each.

\textbf{Example execution:}

\textit{User Query:} ``What's the weather in New York City?''

\textbf{Generated Candidates (temperatures 0.3--1.0):}
\begin{enumerate}
    \item \texttt{get\_weather(location="New York")}
    \item \texttt{get\_weather(location="NYC", temp\_unit="C")}
    \item \texttt{get\_weather(location="New York", temp\_unit="fahrenheit")}
    \item \texttt{get\_weather("New York")}
    \item \texttt{get\_weather(location="NY")}
\end{enumerate}

\textbf{o3-mini scores:}
\begin{itemize}
    \item Candidate 1: 0.8 (missing temperature unit but acceptable)
    \item Candidate 2: 0.3 (uses abbreviation ``C'' instead of full unit name)
    \item Candidate 3: \textbf{0.9} (complete, proper formatting, correct unit)
    \item Candidate 4: 0.6 (missing keyword argument)
    \item Candidate 5: 0.7 (abbreviation ``NY'' less precise than full name)
\end{itemize}

\textit{Result:} Selects Candidate 3 with highest score 0.9.

\subsubsection{Best-of-N Selection (s5): Direct Selection}

The selector mechanism operates similarly to grading but picks the best candidate without explicit scoring.

\textbf{Example execution:}

\textit{User Query:} ``What's the weather in New York City?''

The tool-calling agent generates 5 candidates (same as grading example above).

\textbf{o3-mini evaluates and selects:}

\textit{``Candidate 3 is best: it uses the full city name, includes proper keyword arguments, and specifies the appropriate temperature unit for US locations.''}

\textit{Result:} Selects Candidate 3 directly based on qualitative evaluation.

\subsection{Reviewer Prompt Examples}
\label{sec:appendix-prompts}

We present key versions of reviewer prompts showing the evolution from simple instructions to GEPA-optimized policies. Selector and grader prompts minimally adapt these with their own output sections.

\subsubsection{v1: Baseline Prompt}
\begin{quote}
\small
\raggedright
You are evaluating an assistant's response for correctness, considering the full conversational context.

\#\# Output

Evaluate the assistant's response candidate. Output your evaluation in the following format:

\{output\_start\_tag\}\\
\{\\
\hspace{1em}reasoning: string, (Detailed evaluation reasoning)\\
\hspace{1em}message: string, (Brief explanation of why the response is correct or incorrect)\\
\hspace{1em}error: boolean (Whether response is erroneous or contextually inappropriate)\\
\}\\
\{output\_end\_tag\}

Your response must be a valid JSON object and must be wrapped between \{output\_start\_tag\} and \{output\_end\_tag\} tags.
\end{quote}

\subsubsection{v1-1: With Critical Guideline}
Adds the key insight discovered through error analysis:
\begin{quote}
\small
\raggedright
You are evaluating an assistant's response for correctness, considering the full conversational context.

\textbf{[CRITICAL] Tool-only responses are complete.} DO NOT MARK TOOL-ONLY RESPONSES AS INCOMPLETE ON THE BASIS OF LACKING *USER-FACING ANSWER*, *FOLLOW-UP EXPLANATION*, OR *FINAL RESULTS PRESENTATION* IN THE SAME RESPONSE. Tool call is a standalone step. Marking correct tool-calling responses as incomplete for these reasons is wrong. Focus instead on whether their actual tool calls are correct.

[Output section same as v1]
\end{quote}

\subsubsection{v2-bfcl: Manually Optimized for BFCL}
The baseline manually-engineered prompt (358 tokens).
\begin{quote}
\small
\raggedright
You are evaluating an assistant's response for correctness, considering the full conversational context.

\#\# Criteria

\#\#\# Request Fulfillment\\
- Is the response necessary and reasonably sufficient given the conversational context and the user request?

\#\#\# Tool Call Correctness *(if response invokes tool calls)*\\
- Are the selected tools appropriate?\\
- Any syntax errors relative to the tool doc? Any type errors (e.g., float vs. integer)?\\
- Are argument assignments correct?\\
- \textbf{Accept sensible defaults}: If the parameter values are not explicitly specified in the user request, using default arguments is acceptable.

\#\#\# Other Guidelines\\
- \textbf{Don't be pedantic}: Take a charitable interpretation of the assistant's response. Critique logical failures, NOT surface-level features like tone, style, or minor inefficiencies.\\
- \textbf{No external facts}: Base judgments solely on the conversation, tool doc, and other available information.\\
- \textbf{Binary solvability}: Tasks are designed to be either reasonably solvable or unsolvable.\\
- \textbf{Unsolvability}: If there is no way to reasonably address the request, recognize this as an unsolvable task rather than an assistant error.

[Output section same as v1]
\end{quote}

\subsubsection{v2-tau: Manually Optimized for \texorpdfstring{$\tau^2$-Bench}{Tau2-Bench}}
\begin{quote}
\small
\raggedright
You are evaluating an assistant's response for correctness, considering the full conversational context.

\#\# Evaluation Criteria

\#\#\# Context Awareness (CRITICAL)\\
- Does the response logically follow given the entire conversational context?\\
- If the task explicitly states constraints, policies, or rules for assistant actions, does the response apply them correctly?\\
- If the task explicitly states preconditions for assistant actions, does the response verify them before executing actions?

\#\#\# Request Fulfillment (CRITICAL)\\
- Does the response make a sensible progress towards the complete fulfillment of the user's last request?\\
- Warning: The response might represent one step in an ongoing multi-step request handling. Partial completion is not erroneous if it represents reasonable logical progress.

\#\#\# Logical Consistency\\
- If the assistant communicates tool execution outcomes: Do they match what the tools actually returned?\\
- If the assistant quotes its policy: Does it match what the policy actually states?

\#\#\# Tool Call Correctness *(if response includes tool calls)*\\
- Are the selected tools appropriate?\\
- Any syntax errors relative to the tool documentation?\\
- Are argument assignments correct?

[Additional Guidelines and Output section follow]
\end{quote}

\subsubsection{v3-2025-09-16-bfcl-gepa: GEPA-Optimized}
The most comprehensive version (1,599 tokens). Key sections:
\begin{quote}
\small
\raggedright
You are evaluating an assistant's response for correctness, considering the full conversational context and the available tool documentation. Judge technical and logical correctness of tool use and task coverage; do not critique tone or style.

[CRITICAL] Tool-only responses are complete. DO NOT MARK TOOL-ONLY RESPONSES AS INCOMPLETE ON THE BASIS OF LACKING *USER-FACING ANSWER*, *FOLLOW-UP EXPLANATION*, OR *FINAL RESULTS PRESENTATION* IN THE SAME RESPONSE.

Review policy (apply all):\\
- Evidence scope: Base judgments solely on the conversation and the provided tool documentation. Do not fact-check with outside knowledge.\\
- Tool selection, necessity, and relevance: Approve a tool call only if the tool directly matches the user's intent and can produce the requested target output.\\
- Directness and parsimony: Prefer the minimal number of tool calls that directly yield the requested outputs.\\
- Completeness across multi-step or multi-item requests: If the user asks for multiple actions, all such calls must be present with appropriate arguments.\\
- Argument fidelity: Values must faithfully reflect the user's constraints. Prefer canonical/minimal forms.\\
- Thresholds and inequalities: Unless the tool documentation explicitly specifies strict inequality behavior, treat threshold-like parameters as inclusive bounds.\\
- Units and scales: For rates/percentages, prefer normalized proportions (e.g., 0.03 for 3\%) unless the tool explicitly requires 0--100.\\
- Error criteria (mark error=true when any apply): Wrong or unnecessary tool used; missing required parameters; values that contradict user constraints; mis-scaled units; fabricating missing required values.

[Full prompt available in supplementary materials]
\end{quote}

\subsection{Latency Analysis}
\label{sec:appendix-latency}

We analyze latency overhead of the feedback mechanism across both benchmarks. Table~\ref{tab:latency-stats} presents summary statistics.

\begin{table}[ht]
\centering
\small
\resizebox{\columnwidth}{!}{%
\begin{tabular}{lrrrrr}
\toprule
\textbf{Benchmark} & \textbf{Config} & \textbf{Count} & \textbf{Mean} & \textbf{Median} & \textbf{P95} \\
\midrule
\multirow{2}{*}{BFCL} & Baseline & 10,923 & 1.27s & 0.79s & 3.53s \\
 & +Reviewer & 3,641 & 7.87s & 4.44s & 22.1s \\
\midrule
\multirow{2}{*}{$\tau^2$} & Baseline & 834 & 158.7s & 116.2s & 469.1s \\
 & +Reviewer & 834 & 384.3s & 259.0s & 988.0s \\
\bottomrule
\end{tabular}%
}
\caption{Latency Statistics. BFCL measures per-item latency (seconds). $\tau^2$-Bench measures per-episode duration (seconds).}
\label{tab:latency-stats}
\end{table}

\textbf{Key Observations.} On BFCL (single-turn function calling), the feedback mechanism increases average latency by 6.2×. This substantial increase occurs because the baseline is a single inference call, so reviewer overhead dominates. On $\tau^2$-Bench (multi-turn agent episodes), the mechanism increases average duration by 2.4×. The lower relative impact reflects the multi-turn nature: reviewer overhead is amortized across approximately 40 turns per episode. Figure~\ref{fig:latency-main} shows the latency distributions.

\end{document}